# Advanced robot calibration using partial pose measurements

Alexandr Klimchik
Control System and Industrial Engineering Department
Ecole des Mines de Nantes
Nantes, France
alexandr.klimchik@mines-nantes.fr

Yier Wu
Control System and Industrial Engineering Department
Ecole des Mines de Nantes
Nantes, France
yier.wu@mines-nantes.fr

Stéphane Caro
Research Institute of Communications
and Cybernetics of Nantes
French National Centre for Scientific
Research, Nantes, France
stephane.caro@irccyn.ec-nantes.fr

Benoit Furet
Mechanical Engineering and
Manufacturing Department
University of Nantes
Nantes, France
benoit.furet@univ-nantes.fr

Anatol Pashkevich
Control System and Industrial
Engineering Department
Ecole des Mines de Nantes
Nantes, France
anatol.pashkevich@mines-nantes.fr

*Abstract*—The paper focuses on the calibration of serial industrial robots using partial pose measurements. In contrast to other works, the developed advanced robot calibration technique is suitable for geometrical and elastostatic calibration. The main attention is paid to the model parameters identification accuracy. To reduce the impact of measurement errors, it is proposed to use directly position measurements of several points instead of computing orientation of the end-effector. The proposed approach allows us to avoid the problem of non-homogeneity of the least-square objective, which arises in the classical identification technique with the full-pose information. The developed technique does not require any normalization and can be efficiently applied both for geometric and elastostatic identification. The advantages of a new approach are confirmed by comparison analysis that deals with the efficiency evaluation of different identification strategies. The obtained results have been successfully applied to the elastostatic parameters identification of the industrial robot employed in a machining work-cell for aerospace industry.

## I. Introduction

To achieve the desired accuracy, each industrial robot must go through the calibration procedure, which deals with proper parameter tuning of the mathematical model embedded in the robot controller. Because of its importance, the problem of robot calibration has been in the focus of research community for many years and has been studied from different aspects [1-5]. In spite of this, the issues of the identification accuracy and calibration error reduction have not found enough attention, only limited number of works directly addressed these important problems [3, 5-6]. Generally, two main approaches that allows us to improve the identification accuracy without increasing the number of experiments exist. The first one deals with the preliminary *optimization of the manipulator measurement configurations* (so-called design of calibration experiments). The second approach consists in *enhancing the objective function* to be minimized inside the identification algorithm (in order to minimize impact of the measurement noise). As follows from the literature analysis, the first approach has been considered in a number of papers [7-10], while the second one received less attention of the researches. For this reason, taking into account particularities of the measurement system used in our experiments, this paper focuses on the improvement of the second method, which looks rather promising here.

In robot calibration, there exists a number of techniques that differ in the measurement equipment, the nature of the experimental data (position, orientation, distance, etc.) and in the optimization algorithm that produces the desired parameters [11-14]. At present, the most popular are the so-called open-loop methods that utilise external measurement devices to obtain either full or partial pose of the end-effector (i.e. position and/or orientation) [12-14]. However, it should be noted that the manipulator end-effector orientation cannot be measured directly, so the orientation angles are calculated using positions of several points around the end-effector centre point (TCP). Relevant example can be found in [15], where the end-effector orientation is evaluated using three target points located on the special measurement flange (i.e., three orientation angles are computed from nine Cartesian coordinates provided by the laser tracker). However this approach, which is based on the minimization of the squared sum of the position and orientation residuals, does not allow to minimize the measurement errors impact in the best way (in fact, the position and orientation components of the objective to be minimized are not weighted properly from the statistical point of view). To overcome this difficulty, in this paper it is proposed to use only direct measurement information, i.e. to replace the conventional objective function (squared sum of the position and orientation residuals) by a homogeneous squared sum of position residuals for all measurement points. It is clear that this approach is promising for the identification accuracy improvement, but it requires some revisions of the identification algorithm, which is proposed in this paper.

To address the above defined problem, the remainder of this paper is organized as follows. Section 2 defines the research issue and basic assumptions. In Sections 3, the identification algorithm is presented. Section 4 proposes comparison of the developed identification algorithm with the conventional one. Section 5 presents an application example illustrating benefits of the proposed approach. Finally, Section 6 summarizes the main contributions of the paper.

## II. PROBLEM STATMENT

Let us consider a serial robot whose end-effector location $\mathbf{t} = (\mathbf{p}, \boldsymbol{\varphi})$ (position $\mathbf{p}$ and orientation $\boldsymbol{\varphi}$) is computed using the following vector function

$$\mathbf{t} = g(\mathbf{q}, \boldsymbol{\theta}, \mathbf{\Pi}), \qquad (1)$$

where $g(.)$ defines the manipulator *extended geometric model*, $\mathbf{q}$ is the vector of actuated coordinates, $\boldsymbol{\theta}$ is the vector of robot elastostatic deflections, and the vector of the parameters $\mathbf{\Pi} = \mathbf{\Pi}_0 + \Delta\mathbf{\Pi}$ is presented as the sum of the nominal component $\mathbf{\Pi}_0$ and geometrical errors $\Delta\mathbf{\Pi}$ to be identified via calibration.

In addition to the geometric equation (1), let us consider the *elastostatic model* that allows us to compute the deflections $\boldsymbol{\theta}$ caused by the external loading $\mathbf{F}$ applied to the manipulator end-effector. Relevant equation can be presented in the following form [16]

$$\boldsymbol{\theta} = \mathbf{k}_\theta \cdot \mathbf{J}_\theta^T \cdot \mathbf{F} \qquad (2)$$

where the matrix $\mathbf{J}_\theta = \partial g(\mathbf{q}, \boldsymbol{\theta}, \mathbf{\Pi})/\partial \boldsymbol{\theta}$ is the manipulator Jacobian with respect to the elastostatic deflections $\boldsymbol{\theta}$, and $\mathbf{k}_\theta$ is the manipulator compliance matrix to be identified.

Assuming that the values $\Delta\mathbf{\Pi}$ and $\boldsymbol{\theta}$ are relatively small, equation (1) can be linearized and presented in the form

$$\mathbf{t} = \mathbf{g}_0 + \mathbf{J}_\pi \cdot \Delta\mathbf{\Pi} + \mathbf{J}_\theta \cdot \mathbf{k}_\theta \cdot \mathbf{J}_\theta^T \cdot \mathbf{F} \qquad (3)$$

where the first term $\mathbf{g}_0 = g(\mathbf{q}_i, \mathbf{0}, \mathbf{\Pi}_0)$ corresponds to the nominal geometric model (i.e. to the case when $\boldsymbol{\theta} = \mathbf{0}, \Delta\mathbf{\Pi} = \mathbf{0}$), and the matrix $\mathbf{J}_\pi = \partial g(\mathbf{q}, \boldsymbol{\theta}, \mathbf{\Pi})/\partial\mathbf{\Pi}$ is the manipulator Jacobian with respect to the geometrical parameters $\mathbf{\Pi}$.

The above presented equation (3) is the basic expression for the robot calibration that allows user to obtain the desired geometric and elastostatic parameters $\Delta\mathbf{\Pi}$ and $\mathbf{k}_\theta$. For this purposes a set of experiments are carried out, in which the end-effector locations $\{\mathbf{t}_i\}$ are measured by an external device for several manipulator configurations defined by the vectors of the actuated coordinates $\{\mathbf{q}_i\}$. It is also assumed that the corresponding vectors of the external loading $\{\mathbf{F}_i\}$ are known. It is clear that that corresponding system of linear equations can be solved for $(\Delta\mathbf{\Pi}, \mathbf{k}_\theta)$ if the number of manipulator configurations $m$ is high enough and the vectors $\{\mathbf{q}_i, i = \overline{1, m}\}$ are different to ensure non-singularity of the relevant observation matrix used in the identification procedure. However, there are some difficulties here related to the estimation of the orientation components $\{\boldsymbol{\varphi}_i\}$ of the location vectors $\mathbf{t}_i = (\mathbf{p}_i, \boldsymbol{\varphi}_i)$. There are two main approaches here that are considered below.

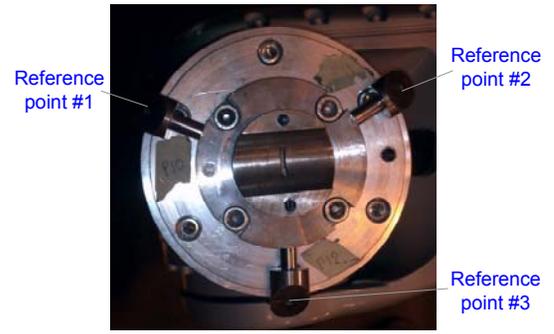

Figure 1. Typical measurement tool used for robot calibration

Usual approach is based on the straightforward utilization of equation (3), where each configuration $\mathbf{q}_i$ produces six scalar equations corresponding to the components of the six-dimensional location vector $\mathbf{t}_i = (p_{xi}, p_{yi}, p_{zi}, \varphi_{xi}, \varphi_{yi}, \varphi_{zi})^T$. Corresponding optimization problem allowing to compute the desired parameters $(\Delta\mathbf{\Pi}, \mathbf{k}_\theta)$ can we written as follows

$$\sum_{i=1}^{m}\left\|\mathbf{t}_i - \mathbf{g}_{0i} - \mathbf{J}_{\pi i}\cdot\Delta\mathbf{\Pi} - \mathbf{J}_{\theta i}\cdot\mathbf{k}_\theta\cdot\mathbf{J}_{\theta i}^T\cdot\mathbf{F}_i\right\|^2 \to \min_{\Delta\mathbf{\Pi}, \mathbf{k}_\theta} \qquad (4)$$

However, in practice, the orientation components $(\varphi_{xi}, \varphi_{yi}, \varphi_{zi})$ cannot be measured directly, so these angles are computed using excessive number of measurements for the same configuration $\mathbf{q}_i$, which produce Cartesian coordinates $\{(p_{xij}, p_{yij}, p_{zij}) \mid j = \overline{1, n}; n \geq 3\}$ for several target points of the measurement tool attached to the manipulator mounting flange (Figure 1). Hence, instead of using $3mn$ scalar equations, that can be theoretically obtained from the measurement data, this conventional approach uses only $6m$ scalar equations for the identification. These may obviously lead to some loss of the parameter estimation accuracy. Another difficulty is related to the definition of the vector norm in equation (4), where the six-dimensional residuals are not homogeneous. It is clear that the position and orientation components must be normalized before computing the squared sum, but it is a non-trivial step affecting the accuracy (in practice, the normalization factors are usually defined intuitively).

To overcome this difficulty it is proposed to reformulate the optimization problem (4) using only data directly available from the measurement system, i.e. the Cartesian coordinates of all reference points $\mathbf{p}_{ij} = (p_{xij}, p_{yij}, p_{zij})^T$ (Figure 1). This idea allows us to obtain homogeneous identification equations where each residual has the same unit (mm, for instance), and the optimization problem is rewritten as follows

$$\sum_{i=1}^{m}\sum_{j=1}^{n}\left\|\mathbf{p}_{ij} - \mathbf{g}_{0ij}^{(p)} - \mathbf{J}_{\pi ij}^{(p)}\cdot\Delta\mathbf{\Pi} - \left[\mathbf{J}_{\theta ij}\cdot\mathbf{k}_{\theta}\cdot\mathbf{J}_{\theta ij}^{T}\cdot\mathbf{F}_{i}\right]^{(p)}\right\|^{2} \to \min_{\Delta\mathbf{\Pi},\mathbf{k}_{\theta}} \quad (5)$$

Here, the superscripts "(p)" indicate the position components (three Cartesian coordinates) of the corresponding location vectors, the index "i" defines the manipulator configuration number, while the index "j" denotes the reference point number. An obvious advantage of this formulation is simplicity in the vector norm definition (conventional Euclidian norm can be applied here reasonably, the normalization is not required) and elimination of the problem of the weighting coefficient selection. In fact, under assumption that the measurement noise is presented as a set of i.i.d. random values (similar for all directions x, y, z and for all measurement configurations), the optimal linear estimator should have equal weights. Besides, there are some potential benefits in the identification accuracy here, since the total number of the scalar equations increases from $6m$ to $3mn$.

To prove advantages of the second approach based on the objective function (5), the following sections will be devoted to the development of the dedicated identification algorithm and the comparison study of the conventional and proposed techniques.

### III. IDENTIFICATION ALGORITHM

Let us assume that the measurement tool has $n$ reference points ($n \geq 3$) that are used to estimate relevant vectors of the Cartesian coordinates $\mathbf{p}_i^j = (p_{xi}^j, p_{yi}^j, p_{zi}^j)^T$ for $m$ manipulator configurations. Using the homogeneous transformation technique, corresponding geometric model (1) can be presented as the matrix product

$$\mathbf{T}_i^j = \mathbf{T}_{base}\cdot\mathbf{T}_{robot}(\mathbf{q}_i, \mathbf{\theta}_i, \mathbf{\Pi})\cdot\mathbf{T}_{tool}^j \quad (6)$$

where the vectors $\mathbf{p}_i^j$ are incorporated in the forth column of $\mathbf{T}_i^j$, the matrix $\mathbf{T}_{base}$ defines the robot base location, the matrices $\mathbf{T}_{tool}^j$, $j = 1,n$ describe locations of the reference points that are observed by the measurement system (see Figure 1), and the matrix function $\mathbf{T}_{robot}(\mathbf{q}_i, \mathbf{\theta}_i, \mathbf{\Pi})$ describes the manipulator geometry and depends on the current values of the actuated coordinates $\mathbf{q}$, the robot elastostatic deflections $\mathbf{\theta}$, and the vector of the parameters $\mathbf{\Pi}$ to be estimated.

Taking into account that any homogeneous transformation matrix $\mathbf{T}_a^b$ can be split into the rotational $\mathbf{R}_a^b$ and translational $\mathbf{p}_a^b$ components and presented as

$$\mathbf{T}_a^b = \begin{bmatrix} \mathbf{R}_a^b & \mathbf{p}_a^b \\ \mathbf{0} & 1 \end{bmatrix}, \quad (7)$$

the Cartesian coordinates of the reference points $\mathbf{p}_i^j, j = \overline{1,n}$ corresponding to the configuration $\mathbf{q}_i$ can be expressed in the following form

$$\mathbf{p}_i^j = \mathbf{p}_{base} + \mathbf{R}_{base}\mathbf{p}_{robot}(\mathbf{q}_i, \mathbf{\theta}_i, \mathbf{\Pi}) + \mathbf{R}_{base}\mathbf{R}_{robot}(\mathbf{q}_i, \mathbf{\theta}_i, \mathbf{\Pi})\,\mathbf{p}_{tool}^j \quad (8)$$

This equation should be accompanied by the elastostatic model $\mathbf{\theta}_i = \mathbf{k}_{\theta}\cdot\mathbf{J}_{\theta}^{T}(\mathbf{q}_i, \mathbf{\theta}_i, \mathbf{\Pi})\cdot\mathbf{F}_i$ that allows us to obtain $3mn$ scalar equations for the identification purposes, where the following vectors/matrices are treated as unknowns: $\mathbf{p}_{base}$, $\mathbf{R}_{base}$, $\mathbf{p}_{tool}^j$, $\mathbf{k}_{\theta}$ and $\mathbf{\Pi}$.

To simplify computations, it is proposed to split the identification procedure into two steps. The first one deals with the estimation of $\mathbf{p}_{base}$, $\mathbf{R}_{base}$, $\mathbf{p}_{tool}^j$, which are related to the base and tool transformations (assuming that the manipulator parameters are known). The second step focuses on the estimation of $\mathbf{k}_{\theta}$ and $\mathbf{\Pi}$ under assumption that the base and tool components are already identified. To achieve desired accuracy, these steps are repeated iteratively several times.

**Step 1**. For the first step, taking into account that the errors in the base orientation are relatively small, the matrix $\mathbf{R}_{base}$ is presented in the following form

$$\mathbf{R}_{base} = [\sim \mathbf{r}_{base}] + \mathbf{I} \quad (9)$$

where $\mathbf{I}$ is $3\times 3$ identity matrix, vector $\mathbf{r}_{base}$ includes the deviations in the base orientation, and the operator $"[\sim]"$ transforms the vector $\mathbf{r} = (r_x, r_y, r_z)^T$ in the skew symmetric matrix as

$$[\sim \mathbf{r}] = \begin{bmatrix} 0 & -r_z & r_y \\ r_z & 0 & -r_x \\ -r_y & r_x & 0 \end{bmatrix} \quad (10)$$

This leads to the following presentation of equation (8)

$$\mathbf{p}_i^j = \mathbf{p}_{base} + \mathbf{p}_{robot}^i - \mathbf{p}_{robot}^i[\sim \mathbf{r}_{base}] + \mathbf{R}_{robot}^i\mathbf{u}_{tool}^j \quad (11)$$

that can also be rewritten in a matrix form as

$$\mathbf{p}_i^j = \mathbf{p}_{robot}^i + \begin{bmatrix} \mathbf{I} & | & [\sim \mathbf{p}_{robot}^i]^T & | & \mathbf{R}_{robot}^i \end{bmatrix} \begin{bmatrix} \mathbf{p}_{base} \\ \mathbf{r}_{base} \\ \mathbf{u}_{tool}^j \end{bmatrix} \quad (12)$$

where $\mathbf{p}_{robot}^i$ and $\mathbf{R}_{robot}^i$ are defined as follows

$$\mathbf{p}_{robot}^i = \mathbf{p}_{robot}(\mathbf{q}_i, \mathbf{\theta}_i, \mathbf{\Pi}); \qquad \mathbf{R}_{robot}^i = \mathbf{R}_{robot}(\mathbf{q}_i, \mathbf{\theta}_i, \mathbf{\Pi}) \quad (13)$$

and

$$\mathbf{u}_{tool}^j = \mathbf{R}_{base}\mathbf{p}_{tool}^j \quad (14)$$

Here the vectors $\mathbf{p}_{base}$, $\mathbf{r}_{base}$ and $\mathbf{u}_{tool}^j, j = \overline{1,n}$ are treated as unknowns.

Applying to the linear system (12) the least-square technique, the desired vectors defining the base and tool transformations can be expressed as follows

$$\left[\mathbf{p}_{base}; \mathbf{r}_{base}; \mathbf{u}_{tool}^1; ... \mathbf{u}_{tool}^n\right] = \left(\sum_{i=1}^{m} \mathbf{A}_i^{j\,T} \mathbf{A}_i^j\right)^{-1} \left(\sum_{i=1}^{m} \mathbf{A}_i^{j\,T} \Delta \mathbf{p}_i\right) \quad (15)$$

where

$$\mathbf{A}_i^j = \begin{bmatrix} \mathbf{I} & \left[\sim \mathbf{p}_{robot}^i\right]^T & \mathbf{R}_{robot}^i & \mathbf{0} & ... & \mathbf{0} \\ \mathbf{I} & \left[\sim \mathbf{p}_{robot}^i\right]^T & \mathbf{0} & \mathbf{R}_{robot}^i & ... & \mathbf{0} \\ ... & ... & ... & ... & ... & ... \\ \mathbf{I} & \left[\sim \mathbf{p}_{robot}^i\right]^T & \mathbf{0} & \mathbf{0} & ... & \mathbf{R}_{robot}^i \end{bmatrix} \quad (16)$$

and the residuals are integrated in a single vector $\Delta \mathbf{p}_i = \left[\Delta \mathbf{p}_i^1; ...; \Delta \mathbf{p}_i^n\right]$. Finally, the variables defining the location to the reference points are computed using expression (14) as $\mathbf{p}_{tool}^j = \mathbf{R}_{base}^T \cdot \mathbf{u}_{tool}^j$. This allows us to find the homogeneous transformation matrices $\mathbf{T}_{base}$ and $\mathbf{T}_{tool}^j$ that are contained in expression (6).

**Step 2.** On the second step, the remaining parameters $\mathbf{\Pi}$ and $\mathbf{k}_\theta$, which define the manipulator geometry and the elastostatic properties, are estimated. For this purpose, the principal system (6) is linearized and rewritten in the form

$$\mathbf{p}_i^j = \mathbf{p}_{robot}^i + \mathbf{J}_{\pi i}^{j(p)} \cdot \Delta \mathbf{\Pi} + \mathbf{A}_{\theta i}^{j(p)} \cdot \mathbf{\chi} \quad (17)$$

where the subscript "$(p)$" denotes the positional components (three first rows) of the corresponding matrices, $\Delta \mathbf{\Pi}$ is the vector of geometrical errors, the vector $\mathbf{\chi}$ collect the unknowns that corresponds to all non-zero elements (by definition) of the compliance matrix $\mathbf{k}_\theta$ (structure of the matrix $\mathbf{k}_\theta$ defines in accordance with the assumptions of links\joins compliance before the identification), the matrix $\mathbf{J}_{\pi i}^j$ is the geometric Jacobian computed for the configuration $\mathbf{q}_i$ with respect to the reference point $j$, and $\mathbf{A}_{\theta i}^{j(p)}$ is derived by relevant transformation of the last term of equation (3) into the vector form:

$$\mathbf{A}_i^j = \left[\mathbf{J}_{1i}^j \mathbf{J}_{1i}^{j\,T} \mathbf{F}_i, ..., \mathbf{J}_{ni}^j \mathbf{J}_{ni}^{j\,T} \mathbf{F}_i\right] \quad (18)$$

In the last expression, $\mathbf{J}_{1i}^j, ..., \mathbf{J}_{ni}^j$ denote the vector-columns obtained by splitting of the geometric Jacobian $\mathbf{J}_{\pi i}^j$. For the computational convenience expression (17) can be presented in the matrix form

$$\Delta \mathbf{p}_i^j = \left[\mathbf{J}_i^{j(p)}, \mathbf{A}_i^{j(p)}\right] \cdot \begin{bmatrix} \Delta \mathbf{\Pi} \\ \mathbf{\chi} \end{bmatrix} \quad (19)$$

where $\Delta \mathbf{p}_i^j = \mathbf{p}_i^j - \mathbf{p}_{robot}^i$ is the residual vector corresponding to the $j$th reference point for the $i$th manipulator configuration.

Appling to this system the least-square technique, the desired vectors $\Delta \mathbf{\Pi}$, $\mathbf{\chi}$, defining the manipulator geometric and elastostatic properties, can be expressed as

$$\begin{bmatrix} \Delta \mathbf{\Pi} \\ \mathbf{\chi} \end{bmatrix} = \left(\sum_{i=1}^{m} \sum_{j=1}^{n} \mathbf{B}_i^{j(p)T} \mathbf{B}_i^{j(p)}\right)^{-1} \left(\sum_{i=1}^{m} \sum_{j=1}^{n} \mathbf{B}_i^{j(p)T} \Delta \mathbf{p}_i^j\right) \quad (20)$$

where $\mathbf{B}_i^{j(p)} = \left[\mathbf{J}_i^{j(p)}, \mathbf{A}_i^{j(p)}\right]$.

It should be noted that, to achieve the desired accuracy, the steps 1 and 2 should be repeated iteratively.

## IV. COMPARISON ANALYSIS

To illustrate the efficiency of the proposed technique, let us compare its accuracy with the conventional one that operates with the full pose information. Their distinctions and particularities can be described as follows:

**Approach #1** (*conventional*): The identification is based on the full pose information $\mathbf{t}_0 = [\mathbf{p}_0, \mathbf{\varphi}_0]$, where both position $\mathbf{p}_0$ and orientation $\mathbf{\varphi}_0$ vectors are computed from the Cartesian coordinates of three reference points $\{\mathbf{p}_1, \mathbf{p}_2, \mathbf{p}_3\}$ located on the manipulator end-effector (see Figure 1).

**Approach #2** (*proposed*): The identification is based on the partial pose information, where three measurement points $\{\mathbf{p}_1, \mathbf{p}_2, \mathbf{p}_3\}$ are directly included in the objective function to be minimized by the identification algorithm.

Let us consider a 3 d.o.f. serial manipulator whose geometry is described by the following equations

$$\begin{aligned} x &= (l_2 \cos q_2 + l_3 \cos(q_2 + q_3)) \cos q_1 \\ y &= (l_2 \cos q_2 + l_3 \cos(q_2 + q_3)) \sin q_1 \\ z &= l_1 + l_2 \sin q_2 + l_3 \sin(q_2 + q_3) \\ \varphi_x &= 0; \quad \varphi_y = q_2 + q_3; \quad \varphi_z = q_1 \end{aligned} \quad (21)$$

where $q_1, q_2, q_3$ are the actuator coordinates and $l_1, l_2, l_3$ are the link lengths to be identified. The manipulator kinematics and procedure of the input data preparation for the approaches #1 and #2 are presented in Figure 2.

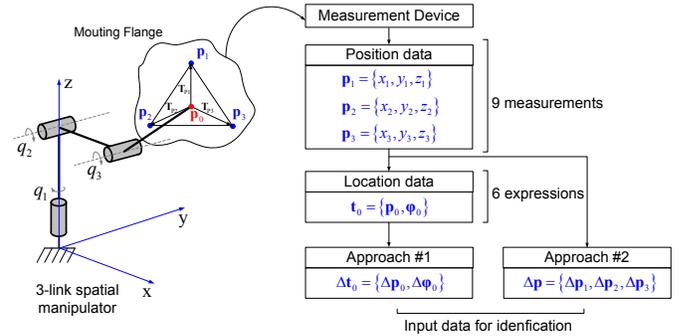

Figure 2. Input data for two identification approaches

For simulation study, the following manipulator parameters have been assigned: $l_1 = 1\,\mathrm{m}$, $l_2 = 0.8\,\mathrm{m}$, $l_3 = 0.6\,\mathrm{m}$. It was assumed that the geometric errors in the link lengths and the actuated joint offsets are respectively $\Delta l_1 = 3.0\,\mathrm{mm}$, $\Delta l_2 = 2.0\,\mathrm{mm}$, $\Delta l_3 = 5.0\,\mathrm{mm}$ and $\Delta q_1 = 1.0\,\mathrm{deg}$, $\Delta q_2 = 0.5\,\mathrm{deg}$, $\Delta q_3 = 2.0\,\mathrm{deg}$. Besides, it was also assumed that the measurements errors are i.i.d. random variables with the standard deviation $\sigma = 0.01\,\mathrm{mm}$ (which perfectly corresponds to the precision of the equipment used for the experimental validations). The desired parameters were estimated using three measurement configurations, which were generated randomly. To obtain reliable statistics, the calibration experiments have been repeated 1000 times.

Simulation results are summarized in Table I, which presents the standard deviations for the estimates of the desired parameters (corresponding mean values of the estimates are equal to the assigned values with the high accuracy). As follows from these results, the proposed approach ensures the accuracy improvement in the estimation of the link length deviations $\Delta l_i$ by the factor of 2.25 ... 3.83, while the accuracy improvement for the joint offsets estimations $\Delta q_i$ is slightly less, up to 3.33. This confirms advantages of the proposed approach, but it should be mentioned that these numbers are obtained for particular set of the measurement configurations and particular normalization factor utilized in the approach #1. However, as follows from our study, approach #2 always provides essentially better results.

TABLE I. IDENTIFICATION ACCURACY FOR DIFFERENT IDENTIFICATION APPROACHES

| Parameter | Standard deviation | | Improvement factor |
|---|---|---|---|
| | *Approach #1* | *Approach #2* | |
| $\Delta l_1$ | 0.069 mm | 0.018 mm | 3.83 |
| $\Delta l_2$ | 0.019 mm | 0.006 mm | 3.17 |
| $\Delta l_3$ | 0.009 mm | 0.004 mm | 2.25 |
| $\Delta q_1$ | 0.187 mdeg | 0.185 mdeg | 1.01 |
| $\Delta q_2$ | 3.742 mdeg | 1.123 mdeg | 3.33 |
| $\Delta q_3$ | 1.432 mdeg | 0.866 mdeg | 1.65 |

## V. APPLICATION EXAMPLE

The developed identification technique has been applied to the elastostatic calibration of industrial robot KR-270 (Figure 3). To take into account the influence of the gravity compensator (which creates the closed-loop) and to apply the virtual joint modeling approach developed for strictly serial robot [16], an equivalent non-linear virtual spring is used (its stiffness depends on the joint variable $q_2$). However, to implement this idea, it is reasonable to consider a set of the compliance coefficients $\{\chi_{2i}, i=1,2,...\}$ corresponding to a number of different joint angles $\{q_{2i}, i=1,2,...\}$ that cover relevant joint limits. This yields the extended set of the elastostatic parameters $\chi_e = [(\chi_{21}, \chi_{22}...), \chi_3, ..., \chi_6]$ to be identified and leads to the linear system of the identification equations similar to those considered in Section III.

To find the measurement configurations that ensure the best identification accuracy, the design of experiments technique has been applied, which is based on the dedicated industry-oriented performance measure proposed in our previous work [17]. This yielded 15 optimal measurement configurations with five different angles $q_2$ that are distributed between the joint limits almost uniformly. These optimal configurations have been obtained taking into account physical constraints that are related to the joint limits and the possibility to apply the gravity force (work-cell obstacles and safety reasons). The results of the calibration experiment design are summarized in Table 2.

At the measurement step, the manipulator was sequentially moved from one configuration to another, where the external loading 250 kg was applied to the special end-effector presented in Figure 4 (it allowed us to generate both external forces and torques). Corresponding experiment setup is shown in Figure 3. To measure the reference point Cartesian coordinates, the laser tracker system Leica AT901 was used. To evaluate manipulator elastostatic deflections, the reference point coordinates have been measured twice, before and after application of the external loading.

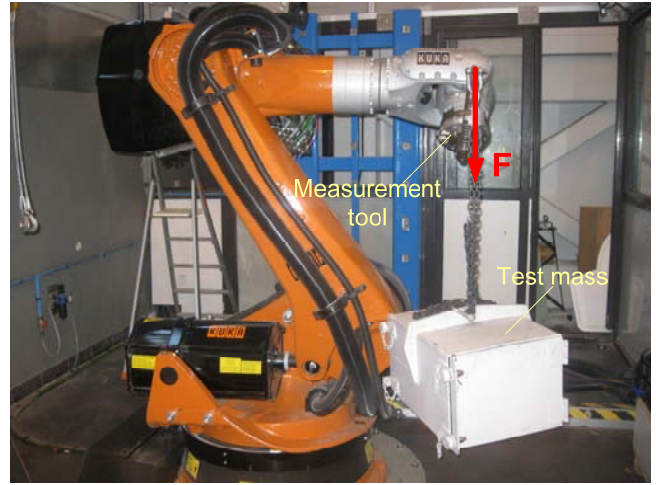

Figure 3. Experimental setup for the identification of the elastostatic parameters

TABLE II. OPTIMAL MEASUREMENT CONFIGURATIONS

| Joint angles, [deg] | | | | | |
|---|---|---|---|---|---|
| $q_1$ | $q_2$ | $q_3$ | $q_4$ | $q_5$ | $q_6$ |
| 79.20 | | -5.57 | 51.00 | -97.52 | -91.67 |
| 63.00 | -0.01 | -12.22 | -56.49 | 41.42 | 150.55 |
| 63.00 | | -47.98 | -70.04 | -61.55 | 177.16 |
| 95.00 | | 33.00 | 129.69 | -98.10 | 90.57 |
| 95.00 | -25.24 | -107.01 | 109.95 | -61.19 | 174.21 |
| 105.00 | | 14.30 | 55.21 | 41.26 | -152.97 |
| 56.60 | | 44.54 | -55.11 | 41.90 | 152.06 |
| 56.60 | -56.9 | 64.73 | -129.65 | -98.260 | -90.55 |
| 144.80 | | 104.49 | -69.41 | 61.67 | -6.33 |
| -41.00 | | -91.68 | 55.12 | 41.53 | -152.48 |
| -143.00 | -99.85 | -32.64 | 110.31 | -61.47 | -6.29 |
| -143.00 | | -72.01 | 129.65 | -98.09 | 90.82 |
| 133.00 | | 147.68 | 129.64 | -97.90 | 90.99 |
| -60.00 | -140 | 7.59 | -110.09 | -61.36 | -174.09 |
| -60.00 | | -52.00 | -124.89 | -41.62 | 27.78 |

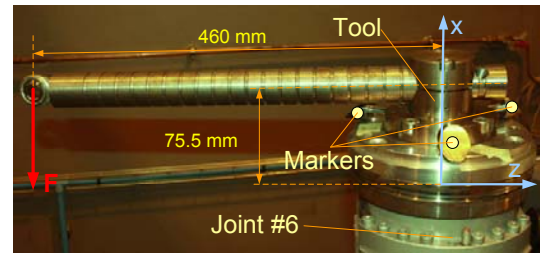

Figure 4. Measurement tool with three markers on the end-effector used for elastostatic calibration experiments

Using these measurement data, the two-step identification procedure has been applied (see Section III). On the first step, the tool and base transformations has been computed; corresponding results are presented in Table 3. On the second step, they have been used for the identification of the elastostatic parameters, which are presented in Table 4. Here values $\chi_{21},...,\chi_{25}$ correspond to five different aggregated compliances of the second actuated joint (which take into account impact of gravity compensator and are obtained for different angles of $q_2$) and $\chi_3,...,\chi_6$ are the compliances of actuated joints 3...6 respectively. These parameters have been further used for the compliance error compensation for the robotic based milling of a aircraft parts, where essential improvement of the precision has been achieved.

TABLE III. IDENTIFICATION RESULTS FOR BASE AND TOOL TRANSFORMATIONS

| $\mathbf{p}_{base}$, mm | $\mathbf{r}_{base}$, mrad | $\mathbf{p}_{tool}^1$, mm | $\mathbf{p}_{tool}^2$, mm | $\mathbf{p}_{tool}^3$, mm |
|---|---|---|---|---|
| $\begin{bmatrix} -34.4 \\ -31.9 \\ -97.8 \end{bmatrix}$ | $\begin{bmatrix} 52.8 \\ 2.2 \\ -15.6 \end{bmatrix}$ | $\begin{bmatrix} 279.2 \\ -16.4 \\ -91.9 \end{bmatrix}$ | $\begin{bmatrix} 279.2 \\ -25.2 \\ 96.1 \end{bmatrix}$ | $\begin{bmatrix} 281.8 \\ 130.5 \\ 5.6 \end{bmatrix}$ |

TABLE IV. IDENTIFIED VALUES OF MANIPULATOR ELASTOSTATIC PARAMETERS

| $\chi_i$ | Values, [rad/N×mm] |
|---|---|
| $\chi_{21}$ | 0.287 ±0.0003 |
| $\chi_{22}$ | 0.277 ±0.0004 |
| $\chi_{23}$ | 0.302 ±0.0005 |
| $\chi_{24}$ | 0.293 ±0.0010 |
| $\chi_{25}$ | 0.246 ±0.0007 |
| $\chi_3$ | 0.416 ±0.0011 |
| $\chi_4$ | 2.786 ±0.0071 |
| $\chi_5$ | 3.483 ±0.0120 |
| $\chi_6$ | 2.074 ±0.0267 |

## VI. CONCLUSIONS

The paper presents the advanced robot calibration technique that is based on the partial pose information, without explicit computation of the end-effector orientation. In contrast to previous works, the proposed technique uses the Cartesian coordinates measurements only, but several reference points of the end-effector are used. The proposed approach allows us to avoid the problem of non-homogeneity of the least-square objective, which arises in the conventional identification technique, where the full-pose information (both position and orientation) is considered. The technique does not require any normalization of the measurement data and can be efficiently applied to both geometric and elastostatic calibration. As follows from the simulation analysis presented in the paper, the developed technique essentially improves the identification accuracy compared to the conventional one. The obtained theoretical results have been successfully applied for the elastostatic parameters identification of an industrial robot used in aerospace industry.

ACKNOWLEDGMENT

The work presented in this paper was partially funded by the ANR, France (Projects ANR-2010-SEGI-003-02-COROUSSO and ROBOTEX). The authors also thank research engineers Fabien Truchet and Joachim Marais for their great help with the experiments.